# A Comprehensive Performance Comparison of Traditional and Ensemble Machine Learning Models for Online Fraud Detection

Ganesh Khekare*
School of Computer Science and Engineering
Vellore Institute of Technology
Vellore, India
khekare.123@gmail.com*

Shivam Sunda
School of Computer Science and Engineering
Vellore Institute of Technology
Vellore, India
shivamsunda10@gmail.com

Yash Bothra
School of Computer Science and Engineering
Vellore Institute of Technology
Vellore, India
yash.bothra27@gmail.com

*Abstract*—In the era of the digitally driven economy, where there has been an exponential surge in digital payment systems and other online activities, various forms of fraudulent activities have accompanied the digital growth, out of which credit card fraud has become an increasingly significant threat. To deal with this, real-time fraud detection is essential for financial security but remains challenging due to high transaction volumes and the complexity of modern fraud patterns. This study presents a comprehensive performance comparison between traditional machine learning models like Random Forest, SVM, Logistic Regression, XGBoost, and ensemble methods like Stacking and Voting Classifier for detecting credit card fraud on a heavily imbalanced public dataset, where the number of fraudulent transactions is 492 out of 284,807 total transactions. Application-specific preprocessing techniques were applied, and the models were evaluated using various performance metrics. The ensemble methods achieved an almost perfect precision of around 0.99, but traditional methods demonstrated superior performance in terms of recall, which highlights the trade-off between false positives and false negatives. The comprehensive comparison reveals distinct performance strengths and limitations for each algorithm, offering insights to guide practitioners in selecting the most effective model for robust fraud detection applications in real-world settings.

## I. INTRODUCTION

Over the years, there has been massive technological development, which has caused great convenience to the financial industry. However, with the growing technology, there has been a rise in online fraud as well, which has caused financial institutions to lose billions of dollars. With the growing rate of fraud, new Machine Learning models are also designed using comprehensive Statistics [1]. The enormous growth of electronic transactions has made it difficult to analyze and classify these frauds. The different ways of financial fraud are discussed: Credit Card Fraud, Mortgage Fraud, Health Care Fraud, and Insurance Fraud. The different Neural Networks that can be deployed to detect such fraud are the Graph Attention Method and the Graph Convolution Method [2]. Financial fraud is a threat not only to economic development but national security, as well, because the profits from fraud may be used for terrorism and other malicious activities. Anomaly detection is used as a technique to curb financial fraud, as there is an extreme imbalance of data and high complexity in financial fraud [3]. Unsupervised machine learning techniques are used as financial fraud is increasing rapidly, making it difficult for existing techniques to detect it. Small devices such as skimmers or shimmers are used to steal information or credit and debit cards. Similarly, in Insurance fraud, forged documents are submitted to manipulate the claim process. Therefore, for such fraud, anomaly detection can be used where the anomaly is defined as patterns in data that deviate from the defined characteristics of the normal behavior [4].

Traditional detection methods are often inadequate to counter these evolving threats. Machine Learning presents a powerful alternative by analyzing transactional patterns to detect anomalies that may signal fraud. This research focuses on the advantages of machine learning systems like SVM, Random Forest, Logistic Regression, XGBoost, and ensemble methods like Stacking and Voting Classifiers for credit card fraud identification [5]. These systems are chosen for their robust performance in classification tasks, managing multivariate datasets, and generalizability, especially when dealing with noisy or imbalanced datasets. One of the significant difficulties in the identification of fraudulent transactions is the imbalance in the dataset, where a larger chunk of data represents legitimate interactions [6]. Because of this issue, there can be a bias towards predicting legitimate transactions. To address this issue, various data balancing methods can be applied along with comprehensive evaluation metrics to assess model performance. In this study, the dataset goes through preprocessing levels like attribute scaling, handling null entries, and balancing the class distribution. Then, each system is trained using the information after preprocessing and evaluated using important metrics like F1-score, accuracy, AUC-ROC, precision, recall, and precision-recall curve. After this, the various models are compared to identify a superior balance between recall and precision, getting a reliable fraud identification model while simultaneously reducing the false negatives and false positives.

## II. LITERATURE REVIEW

In recent history, the increasing complexity of online fraud has necessitated the evolution of more solid and precise fraud identification models. Several machine learning techniques have been developed to enhance the accuracy and efficacy of fraud identification techniques. Among these models, SVM, Logistic Regression, XGBoost, and Random Forest have become particularly popular due to their apparent strengths in handling categorization problems. Ensemble methods can be an altar- native to the single model approach as significant contributions have been made in this field. This study [7] directly explores ensemble methods, including Voting Classifiers, and highlights XGBoost as a high-performing model. On combining models like Random Forests and AdaBoost, the results indicate that it enhances performance, which is particularly relevant to the focus on Stacking and Voting Classifiers. This reinforces the idea that ensemble methods can provide a more general detection capability than individual models [8].

In [9], it is seen that Logistic Regression and decision trees also provide valuable insights in fraud detection. The challenges associated with data imbalance and model overfitting, which are common issues in the making of accurate fraud identification models, are also discussed. By implementing advanced methods such as the JAYA optimization algorithm, the study also demonstrates how to improve the model per- performance in classifying fraudulent and legitimate transactions. Since both Logistic Regression and SVM are sensitive to class imbalance, their performance can be significantly enhanced through effective preprocessing techniques. In [10], when various machine learning models like SVM, Random Forest, and KNN were compared for credit card fraud identification, it was found that both SVM and Random Forest achieved high accuracy. The comparison metrics show that SVM excels in precision, while Random Forest demonstrates superior overall effectiveness because of its potential to identify complex data patterns. The research demonstrates that although SVM is a strong performer in precision, discrete models like Random Forests and XGBoost provide superior performance in practical implementation. The emphasis of feature engineering and leveraging preprocessing techniques to optimize the impact of machine learning algorithms is discussed in [11]. Moreover, the applications of oversampling techniques, such as SMOTE, to manage complex data and improve the accuracy of systems like XGBoost and Random Forest are also discussed. By showing how feature engineering can lead to significant performance improvements, the study supports the inclusion of feature engineering strategies in ensemble methods like Stacking, where multiple models are utilized to optimize prediction accuracy. Similarly, the understanding of feature selection's role in fraud detection can be enhanced by the employment of the Oppositional Cat Swarm Optimization (OCSO) algorithm to select relevant features, which resulted in an impressive accuracy of 99.97%. The results highlight the importance of using optimization methods to enhance system efficiency.

The results of decision trees in fraud detection are seen, which emphasizes the significance of various performance metrics, implying that these are essential aspects to consider when comparing different machine learning models [12]. The focus is on reducing false positives and false negatives, especially as ensemble methods like Voting and Stacking aim to mitigate these issues by aggregating the outputs of various models to improve overall accuracy. The advancements in SVM for fraud detection provide further context for the detection of fraudulent activities. The effectiveness of SVM in processing high-dimensional, multi-variate, and imbalanced datasets addresses the limitations faced by traditional ma- machine learning algorithms. The results suggest that while SVM can deliver high accuracy, integrating it with ensemble techniques like Stacking could further enhance its performance in detecting fraud. The relative assessment of Logistic Regression, SVM, XGBoost, Random Forest, and ensemble systems like stacking and voting reveals that no single model is universally superior. Instead, merging the strengths of various systems through ensemble methods offers the most promising approach to improving fraud detection accuracy, especially when resolving hurdles like data complexity and overfitting, offering valuable insights into model optimization, feature selection, and real-time detection strategies.

Apart from regular machine learning models, efficient Graph Neural Networks (GNNs) should be enabled. However, when GNNs are implemented, there are two main challenges: i) Protecting by carefully doing upcoming deals. ii) The performance of Graph analysis. To address these issues, BRIGHT was developed, which consists of a GNN layout along with a graph-modifying module. The GNN layout is a Lambda Neural Network, whereas the graph modifying module is a two-stage directed graph. It surpassed baselines by 2% concerning mean precision and considerably lowered response time by approximately 7.8x. The labels are assigned to unlabelled data using pseudo labeling, because of which more models are trained based on the classified pseudo labels using the limited labeled data. Autoencoders were used so that the pseudo labelling process could become better. It was observed that semi-supervised learning outperforms simple models like logistic regression and random forests. The importance of the portion of pseudo-labelled attribute clusters and the contribution of the autoencoders was one of the most striking features, and modifications like gradually increasing the lambda while training can be performed on the model. The databases used for this model included Yelp Online Review, Amazon Online Review, and the Elliptic Bitcoin Fraud Dataset.

## III. METHODOLOGY

This section of the research paper describes a structured approach and techniques used to compare the performance of diverse machine learning methods like Random Forest, SVM, Logistic Regression, XGBoost, Stacking Classifier, and Voting Classifier— in identifying debit card fraud. This section explains the details of information preprocessing, the metrics used to evaluate each algorithm, and the implementation of these algorithms. Figure 1 shows before and after the results of balancing the data.

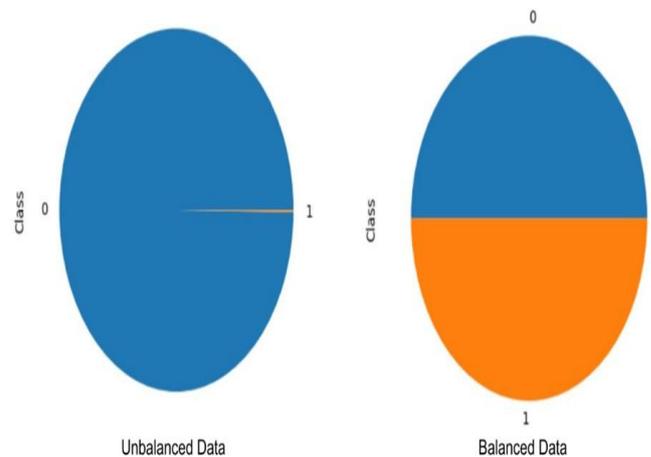

Fig. 1. Before and after balancing the data

The selected models have both traditional and advanced machine learning techniques, each with its unique strengths and weaknesses. Ensemble methods were chosen as it combine the advantages of various machine learning models, which improves performance metrics. A single model may not be able to detect patterns and predict fraud. By using different models, more patterns can be observed, and based on that, reliable predictions can be made.

*A. Data Preprocessing*

This includes the collection and preparation of data. The data used has credit card transactions marked as legitimate or fraudulent. The dataset preparation includes handling outliers, null values, missing values, and normalization so that the system can be efficiently trained on the dataset.

*B. Model Implementation*

- Random Forest: A machine learning model that utilizes more than one decision tree created during training to give a single result.
- SVM: A machine learning algorithm that uses a hyperplane between data points to produce the result.
- Logistic Regression: A numerical system that systems the probability that the provided input is of a particular class.
- XGBoost (Extreme Gradient Boosting): A refined distributed gradient boosting library designed to be portable, flexible, and efficient.
- Stacking Classifier: Merges different machine learning algorithms by using their forecasts as inputs to a final model for better accuracy.
- Voting Classifier: Adds identifications from different machine learning systems (by majority voting probabilities) to predict the last stage.

By practically implementing these machine learning models using a consistent methodology, it would be easier to point out which algorithm is the best suited for debit card fraud identification. Implementing would also shed light on the pros and cons of each system so that practitioners can choose the most suitable method for their specific needs.

*C. Dataset Collection*

The data set utilized for online transactions fraud identification is sourced from Kaggle. It includes the following columns:

- Time: This represents the elapsed time between transactions.
- Amount: Indicates the amount of the transaction represented in US$.V1, V2, V3, ... V28: These columns capture various anonymous features related to the transaction, including aspects like position and type of the transaction.
- Class: Shows whether the transaction is fraudulent or not. This column serves as the target variable for prediction by the machine learning model.

*D. Data Preparation*

In this dataset, V1, V2, V3, ..., and V29 are features of the transactions, but the details cannot be shared because credit card information is sensitive and private and cannot be exposed. The features are converted using PCA (Principal Component Analysis), and all are represented as numerical values. The Time column is in seconds and represents the elapsed time relative to the first transaction (for example, if row 23 has a time of 17 seconds, it indicates that this transaction occurred 17 seconds after the first transaction). The Class column shows whether the interaction is fraudulent (1) or not fraudulent (0). The data includes 284,315 legal transactions (denoted by 0) and 492 fraud entries (which is shown by 1), making it highly unbalanced, as shown in Fig. 1. Running a model on this dataset without adjustment would not yield satisfactory results, so a balanced dataset is required. To balance the dataset, 492 legitimate transactions are randomly selected to match the number of fraudulent transactions. Fig. 2 shows the flowchart for machine learning algorithms.

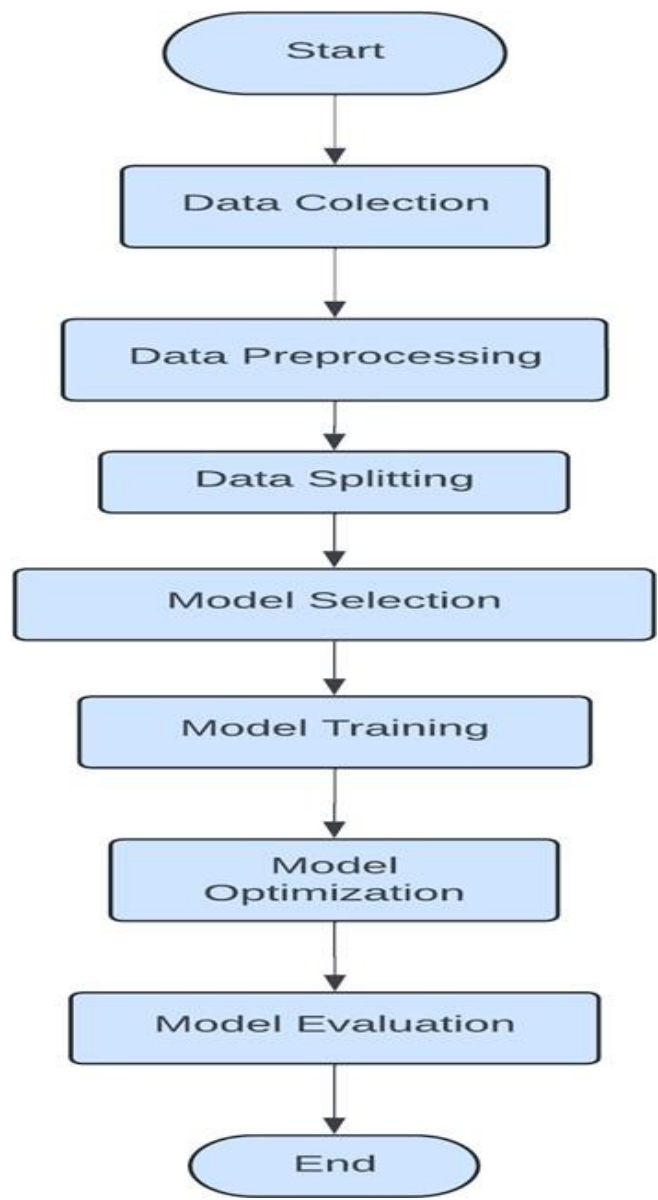

Fig. 2. Flowchart for a machine learning algorithm

*Linear Relationship*: Logistic Regression presupposes a direct correlation among the input characteristics and the log odds of the result. The log-odds represent the log of the chances of the outcome happening multiplied by the reciprocal of the probability of it not happening.

*Logistic Function*: The logistic technique, also known as the sigmoid method, is applied to convert the input features' linear variation into a probability. This function ensures that the result stays between zero and one.

Training the model involves the algorithm determining the coefficients of the linear equation that most accurately represents the data by increasing the chance of observing the data provided. This is commonly achieved through techniques such as gradient descent.

*Forecast*: When given a new input, the model computes the weighted total of the input features, applies the logistic function, and produces a probability. A class label is determined by keeping a threshold, generally keeping at 0.5.

Logistic Regression is beneficial because it is straightforward- ward, easy to understand, and quick. It offers a straightforward probabilistic structure for categorization, simplifying comprehension and application. Furthermore, Logistic Regression is less likely to overfit if the entries of iterations is less than the number of attributes. XGBoost is a high-level gradient boosting library that is highly efficient and scalable. Because of its capacity to effectively manage large datasets and complex models, it is employed for both regression and classification tasks. XGBoost works based on boosting, which consists of creating a sequence of weak learners one after another, with every learner fixing the mistakes of the ones before it. Various machine learning methods like Random Forest, SVM, and XGBoost made their predictions. In the voting classifier, the most optimal predictions were picked from these. In stacking, a model was implemented that learned from all the other models' outputs and made a decision based on that combined output. Python and its libraries were used to build and test the models. The data was first cleaned and normalized. Then the models were trained on the training data and later evaluated. Based on the accuracy, precision, and recall, further fine-tuning was done.

## IV. RESULTS AND DISCUSSION

In this study on fraud identification using different ma-machine learning algorithms, the comparison of several algorithms—Logistic Regression, SVM, XGBoost, Random Forest, Voting Classifier, and Stacking—is based on their performance metrics: accuracy, recall, precision, and F1 score. The objective was to pick out the systems that are most effective at identifying fraudulent entries while minimising false negatives and false positives. Fraud identification is an important thing where both misclassifying fraudulent cases (false negatives) and flagging legal entries as fraudulent (false positives) can have various effects.

After training and implementing the models on the dataset, it was observed that the Voting Classifier, SVM, and Random Forest have the highest accuracy at 0.94, showing their strong ability to classify transactions correctly. Stacking and Logistic Regression were just behind with 0.93 accuracy, while XGBoost had a slightly lower accuracy of 0.91. This shows that all models perform reasonably well in terms of general accuracy, with minor variations, as shown in Figure 3.

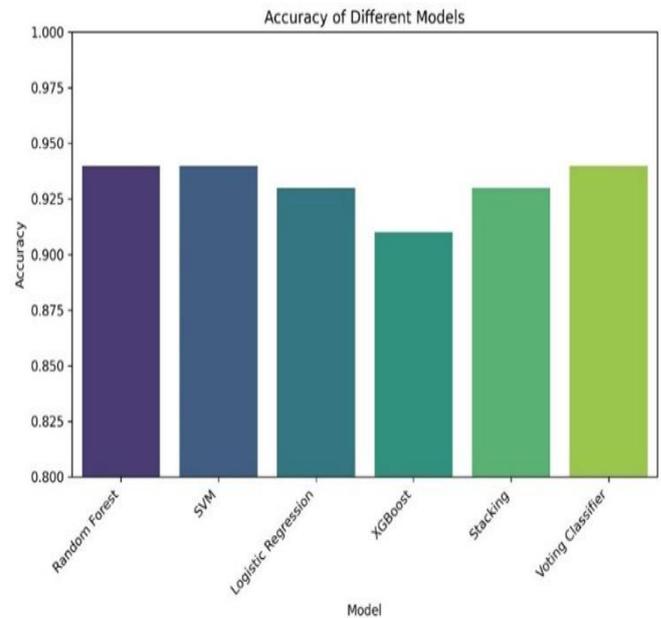

Fig. 3. Comparison of machine learning techniques based on accuracy

When comparing the precision and recall scores of classes 1 and class 0 transactions, the results have shown some key differences. For Precision (Class 1), which measures how well the models correctly detect fraudulent entries while reducing false positives, the Voting Classifier and Stacking models stood out, achieving almost perfect precision at 0.99 and 0.98, as shown in Figure 4. This implies that these models are highly efficient at correctly detecting fraudulent cases without mistakenly flagging too many legitimate transactions. However, in terms of Recall (Class 1), which measures how many actual fraudulent transactions are detected, SVM, Random Forest, and Logistic Regression outperformed the ensemble methods. Their recall values ranged between 0.92 and 0.93, as shown in Figure 5, meaning they are better at capturing more fraudulent cases but may still produce some false positives.

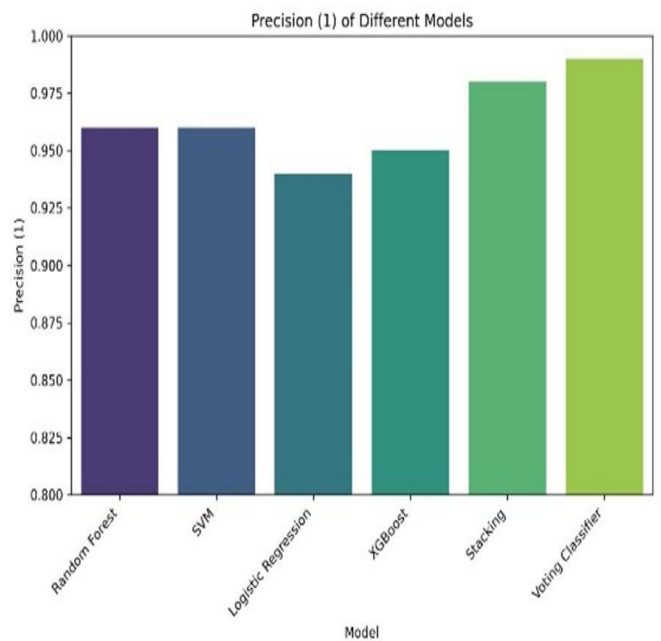

Fig. 4. Comparison of machine learning models based on precision

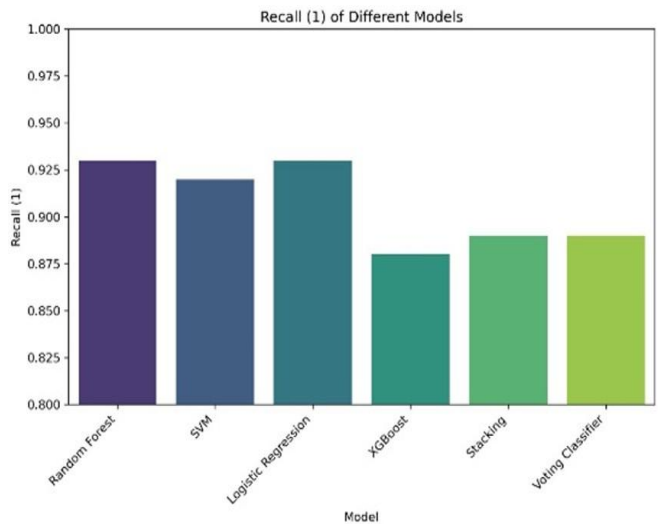

Fig. 5. Comparison of machine learning models based on recall

F1 scores further highlight the link between precision and recall by merging them into a single attribute. In this regard, Voting Classifier, Stacking, and Random Forest performed equally well with F1-scores of 0.93 and above, as shown in Figure 6, suggesting that they offer a good balance between capturing fraudulent cases and minimizing false alarms.

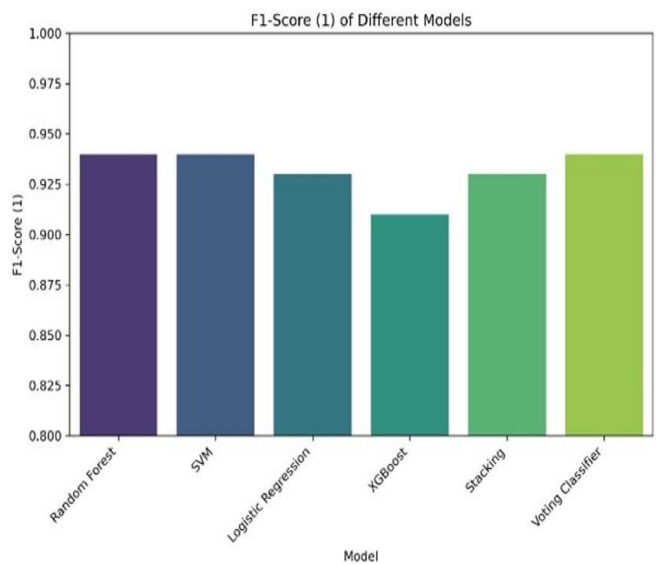

Fig. 6. Comparison of machine learning models based on F1-score

In the comparison of machine learning techniques, while SVM was used for classification, it was excluded from the precision-recall and ROC curve analyses due to issues with its probability estimation. Unlike models such as Logistic Regression, SVM does not natively produce probability estimates; it relies on Platt scaling, which often leads to poorly calibrated probabilities, especially in imbalanced data such as fraud identification. This affects the accuracy of precision-recall and ROC curves, which depend on reliable probabilities to evaluate the model. Performance across different thresholds, including SVM in these analyses, could have provided misleading results. Thus, while SVM was part of the broader classification comparison, it was excluded from these specific performance evaluations to ensure the results remained reliable and consistent across models. The comparative analysis of different algorithms has been carried out based on multiple parameters.

## V. CONCLUSION

The evaluation and comparison of traditional and ensemble machine learning models for online fraud detection showed that a model should be chosen according to specific requirements. Traditional methods like Random Forests and SVM showed better stability between precision and recall when compared to other models. This increases the identification of fraudulent transactions but also increases false positives. Ensemble methods have a high precision, which minimizes false positives; this makes them highly suitable in the case of imbalanced datasets. By analyzing the Precision-Recall curve and the ROC-AUC model, a specific model can be chosen on the basis of trade-offs between minimizing false positives and maximizing fraud detection. Both traditional as well as ensemble methods showed high AUC and F1 scores, demonstrating strong performance metrics. Overall, it was found that ensemble learning methods are more effective in the detection of fraudulent transactions.

## VI. FUTURE SCOPE

Future research could explore better and more advanced techniques for issues related to imbalanced datasets. Instead of simply randomly choosing the values to make the dataset balanced, approaches like SMOTE can be used, which can prove to help enhance the model performance. Having a more balanced dataset can improve the overall robustness of the model. Apart from this, future work should focus on improving the SVM model so that it can be employed for calculating the Precision-Recall Curve and ROC Curve, for which one of the methods could be using Calibrated Classifier CV. Further work should be done on evaluating other algorithms or hybrid approaches, or by incorporating Deep Learning algorithms like ANN (Artificial Neural Networks) or CNN (Convolutional Neural Networks), along with the different Machine Learning algorithms, and using more efficient ensemble methods apart from Stacking and Voting Classifier. Performance evaluation can also be done on the RXT-Next model, which could also be very helpful for online fraud detection. Thus, working on the above-mentioned models may improve detection capabilities and yield more reliable results in real-world applications.